\documentclass[a4paper,conference]{IEEEtran} 
\IEEEoverridecommandlockouts             

\usepackage{graphics} 
\usepackage{graphicx}
\usepackage{hyperref}
\usepackage{url}
\usepackage{scrextend}
\usepackage{amsmath} 
\usepackage{capt-of,etoolbox}
\usepackage{amsfonts}
\usepackage{listings}
\usepackage{xspace}
\usepackage{multirow}
\usepackage{array}
\usepackage[table,xcdraw]{xcolor}
\usepackage{url}
\usepackage{booktabs} 
\usepackage{mathtools}
\usepackage{mathtools, nccmath}

\usepackage{etoolbox}
\makeatletter
\patchcmd{\@makecaption}
  {\scshape}
  {}
  {}
  {}
\makeatother

\DeclarePairedDelimiter\ceil{\lceil}{\rceil}
\DeclarePairedDelimiter\floor{\lfloor}{\rfloor}
\DeclarePairedDelimiter{\nint}\lfloor\rceil

\DeclareGraphicsExtensions{.pdf,.png,.jpg}
 
\usepackage[lmargin=13.1mm, rmargin=13.1mm, tmargin=19.1mm,bmargin=39.9mm]{geometry}

\newcommand{\eg}{e.g. }
\newcommand{\ie}{i.e. }
\newcommand{\algoname}{\text{LOOP}}
\newcommand{\algoacro}{\textbf{L}everaging on a generic \textbf{O}bject detector for \textbf{O}rientation \textbf{P}rediction}
\newcommand{\datasetName}{LOOP Dataset}

\DeclareMathOperator{\Real}{\mathbb{R}}
\DeclareMathOperator{\IntegersP}{\mathbb{Z}^+}
\DeclareMathOperator{\lmpoint}{\mathbf{p}_{lm}}

\newcommand{\nAngles}{k}

\def\BibTeX{{\rm B\kern-.05em{\sc i\kern-.025em b}\kern-.08em T\kern-.1667em\lower.7ex\hbox{E}\kern-.125emX}}

\begin{document}

\makeatletter
\newenvironment{ntp}{%
  \@beginparpenalty\@lowpenalty
  \bfseries\small\textit{Note to Practitioners}\textemdash
  \@endparpenalty\@M}%
{\par}
\makeatother


\title{Effective Deployment of CNNs for 3DoF Pose Estimation and Grasping in Industrial Settings\\
\thanks{This work was supported by the European Union’s Horizon 2020 research and innovation programme under grant agreement No. 870133 as part of the RIA project REMODEL (Robotic tEchnologies for the Manipulation of cOmplex DeformablE Linear objects).}
}

\author{\IEEEauthorblockN{Daniele De Gregorio} \IEEEauthorblockA{EYECAN.ai\\
Bologna, Italy \\
daniele.degregorio@eyecan.ai}
\and
\IEEEauthorblockN{Riccardo Zanella}
\IEEEauthorblockA{
\textit{DEI - University of Bologna}\\
Bologna, Italy \\
riccardo.zanella2@unibo.it}
\and
\IEEEauthorblockN{Gianluca Palli}
\IEEEauthorblockA{
\textit{DEI - University of Bologna}\\
Bologna, Italy \\
gianluca.palli@unibo.it}
\and
\IEEEauthorblockN{Luigi Di Stefano}
\IEEEauthorblockA{
\textit{DISI - University of Bologna}\\
Bologna, Italy \\
luigi.distefano@unibo.it}
\vspace*{-50mm}
}

\maketitle

\begin{abstract}
In this paper we investigate how to effectively deploy deep learning in practical industrial settings, such as robotic grasping applications. When a deep-learning based solution is proposed, usually lacks of any simple method to generate the training data. In the industrial field, where automation is the main goal, not bridging this gap is one of the main reasons why deep learning is not as widespread as it is in the academic world. For this reason, in this work we developed a system composed by a 3-DoF Pose Estimator based on Convolutional Neural Networks (CNNs) and an effective procedure to gather massive amounts of training images in the field with minimal human intervention. By automating the labeling stage, we also obtain very robust systems suitable for production-level usage. An open source implementation of our solution  is provided, alongside with the dataset used for the experimental evaluation.   \end{abstract}


\begin{figure*}
  \centering
\includegraphics[width=0.67\linewidth]{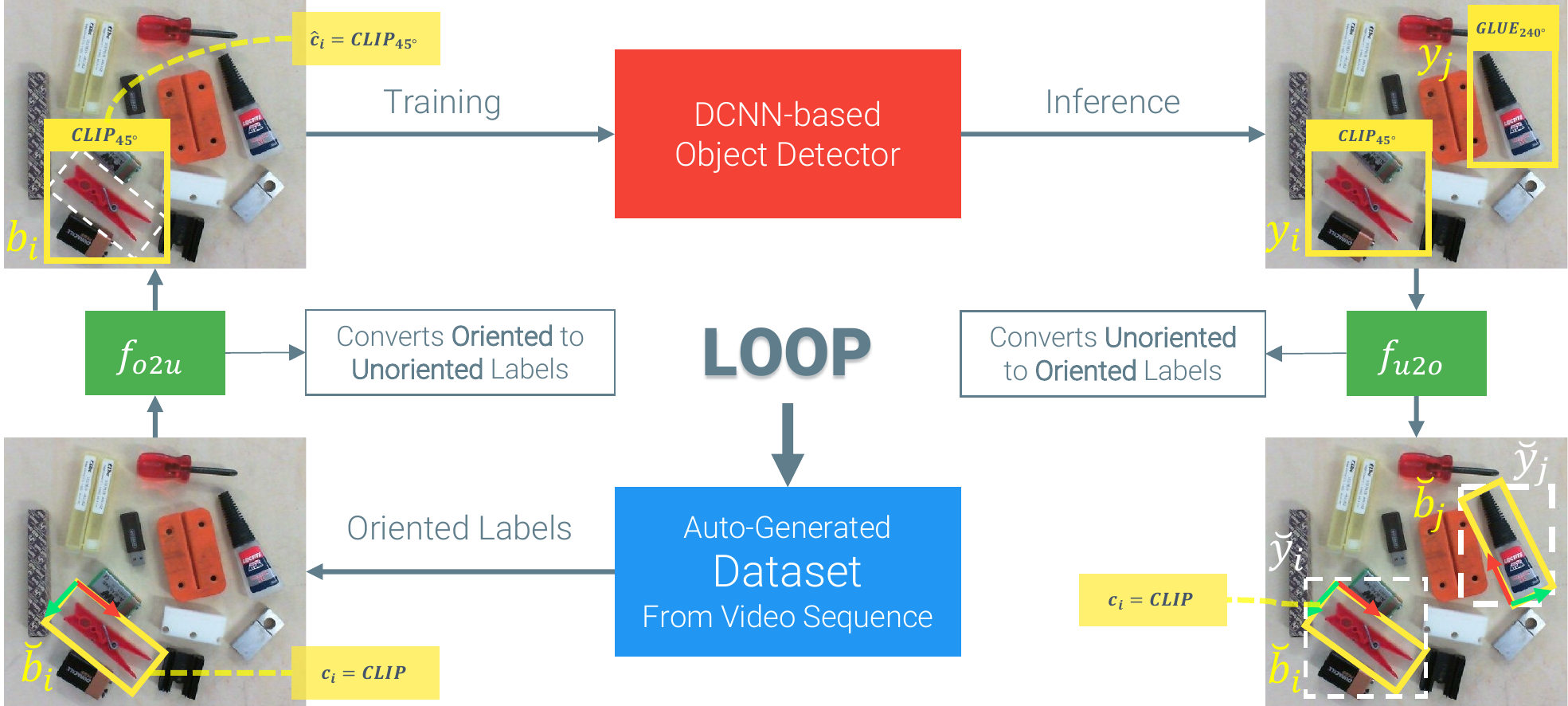}
\caption{The overall pipeline of \algoname{}. The starting point is the creation of a dataset of oriented bounding boxes (oriented labels) with minimal human intervention. The \emph{oriented} labels are converted in \emph{unoriented} labels, by means of the $f_{o2u}$ function , suitable to train a classical object detector, which encodes objects orientation in the classification process. Hence the detector infers \emph{unoriented} predictions ($y_i$), with the same orientation encodings. The \emph{unoriented} predictions are then transformed in \emph{oriented} predictions ($\breve{y}_i$)  by means of the $f_{u2o}$ function. }
\label{fig:teaser}
\vspace*{-3mm}\end{figure*}

\section{Introduction}\label{sec:introduction}

Among visual perception tasks, 2D Object Detection with category-level classification has achieved an effectiveness thanks to  Convolutional Neural Networks (CNNs)  \cite{redmon2016you},\cite{liu2016ssd},\cite{ren2015faster}. Unfortunately, the more complex perception task of the 3D Pose estimation has not experienced the same strengthening, notwithstanding remarkable results \cite{kehl2017ssd},\cite{tekin2018real},\cite{sundermeyer2018implicit} have endorsed CNNs also for this -- more complex -- task. We argue that between the key reasons for this state-of-affairs is the lack of training data: while for 2D Object Detection huge datasets like Pascal VOC \cite{pascal-voc-2012} or COCO \cite{lin2014microsoft} define a reliable testbed for the community, the same can not be said for the 3D counterpart (with the exception of some small datasets \eg like \cite{hinterstoisser2011multimodal},\cite{hodan2017t}).
Obviously, for relevant robotic applications, such as, for instance, a fully automatic  pick-and-place, Pose Estimation is an essential stage of the overall pipeline, and, as stated before, the availability of training data hinders  deployment of CNNs. Thus, the claim of our work is that for a real industrial application it is not sufficient to develop advanced data-driven  models, like a convolutional neural networks, but  -- simultaneously  -- the data sourcing problem should be addressed. Thus,  we  propose to tackle the object detection and  3-DoF pose estimation task by an integrated framework based on CNNs wherein the required labeled training data are  sourced autonomously, \ie with negligible  human intervention. We also show  how the proposed framework enables development of  a fully automated robotic grasping system.

The basic idea of our approach is to take advantage of the known difficulty of CNNs to learn rotational invariant image features. As shown in \cite{zhou2017oriented},\cite{zeiler2014visualizing} these networks redundantly learn multiple representation of sought objects when they exhibit multiple rotation in the training images. As opposed to approaches like \cite{jaderberg2015spatial},\cite{zhou2017oriented}, which try to learn rotation-invariant representation, we leverage on classical CNN-based Object Detectors to formulate the angle estimation task as a classification problem by leading the network to interpret each single object orientation as a stand-alone class (for this reason we name the algorithm \textbf{\algoname{}}: \algoacro{}). As previously mentioned, we endow our approach with an automated dataset generation technique that allows  to label an entire video sequence easily, provided that the sequence features the same type of image acquisitions conditions as those in which the detector is exptected to operate at test time (\eg a quite-planar scene in front of the camera). This approach enables collections of massive amounts of training data which, in turn,  allow the creation of an almost perfect object detector (\ie $\sim 0.99$ \emph{mAP} according to our experiments).

Thus, the key contributions of our paper are:

\begin{itemize}
\item A novel approach to extend a generic CNN-based 2D Object Detector in order to predict oriented bounding boxes.
\item A fast and reliable Labeling pipeline that allows to gather  a labeled Dataset to train the above mentioned extended Object Detector with minimal human intervention (\ie the user has to manually label only the first frame of a video sequence).
\item A real -- proof-of-concept -- pick\&place robotic application based on this approach.
\end{itemize}

\noindent An open source implementation of the proposed method is available online \footnote{
\label{note:code_url}
\url{https://github.com/m4nh/loop}
}. 

\section{Related Works}\label{sec:related}

Nowadays, object detection and 3-DoF Pose Estimation in industrial settings is mainly addressed by classical computer vision approaches based on hand-crafted 2D features, which can effectively represent the orientation of salient local image structures. SIFT \cite{lowe1999object} is one of the most popular 2D feature detector and descriptor by which it is possible to implement a full Object Detection pipeline for textured objects. By matching multiple 2D local features it is possible to estimate the homography (or even a rigid transform) between a model image and the target scene. SIFT can be replaced by other popular alternatives, like SURF, KAZE, ORB, BRISK etc.. \cite{tareen2018comparative}  presents a comprehensive evaluation of the main algorithms for 2D feature matching. As stated before, the aforementioned methods are suitable -- mostly -- for \emph{textured} objects detection, but the same matching pipeline can be deployed replacing them with detectors/descriptor based on geometrical primitives (\eg oriented segments) amenable to texture-less objects. One of the leading \emph{texture-less} object detectors is BOLD \cite{tombari2013bold}, which was then followed by BORDER \cite{chan2016border} (and its extension, referrred to as BIND \cite{chan2017bind}).
A popular alternative to features, instead, is Rotation-Invariant Template Matching like OST \cite{liu2018fast}, OCM \cite{ullah2004using} or Line2D \cite{hinterstoisser2010dominant}. However, a tamplate-based approach may not be the most efficient solution for real-time applications. 

The above-mentioned considerations have lead us to compare \algoname{} mainly with SIFT \cite{lowe1999object} and BOLD \cite{tombari2013bold}, undoubtedly two state-of-the-art approaches in \emph{textured} and \emph{texture-less} object detection, respectively. Our claim is to propose a real-time deep learning alternative able to cope with both textured as well as untextured models and, seamlessly,  over plain and cluttered backgrounds.  As stated in \autoref{sec:introduction}, as we can exploit any generic CNN Object Detector, we investigated here about well established approaches like YOLO \cite{redmon2016you} and SSD\cite{liu2016ssd}, which resolve the 2D object detection and recognition task with a single image analysis  pass,  as well as  Faster R-CNN\cite{ren2015faster}, which instead conceptually splits the detection and recognition parts. We found that YOLOv3 \cite{redmon2018yolov3}, the newest declension of the classical YOLO algorithm, represents a satisfactory test case for our experiments, though is worth pointing out that  \algoname{} is detector-agnostic. 

Another research line  to attain robotic grasping systems concerns  direct estimation of grasping points from images by means of CNNs. One of the most used approach, conceptually similar to our method, is the 2D Rectangular Representation of grasp, as described in \cite{jiang2011efficient}. The authors demonstrated that a 2D representation of grasp is enough to perform a 3D manipulation with a robotic arm. Recent works like \cite{redmon2015real} or \cite{chu2018real} estimate the position and the orientation of these 2D Rectangular Representation of grasp by means of a CNN, as either a regression or classificatiton problem, respectively. However, we believe that  the full 2D Oriented Bounding Box of the sought objects yielded by our approach is a better representation for grasp in planar setting, because it allows to perform both obstacle avoidance as well as model-based grasp points computation.

\section{How does \algoname{} work?}

\begin{figure}[t]
  \centering
  \includegraphics[width=0.85\columnwidth]{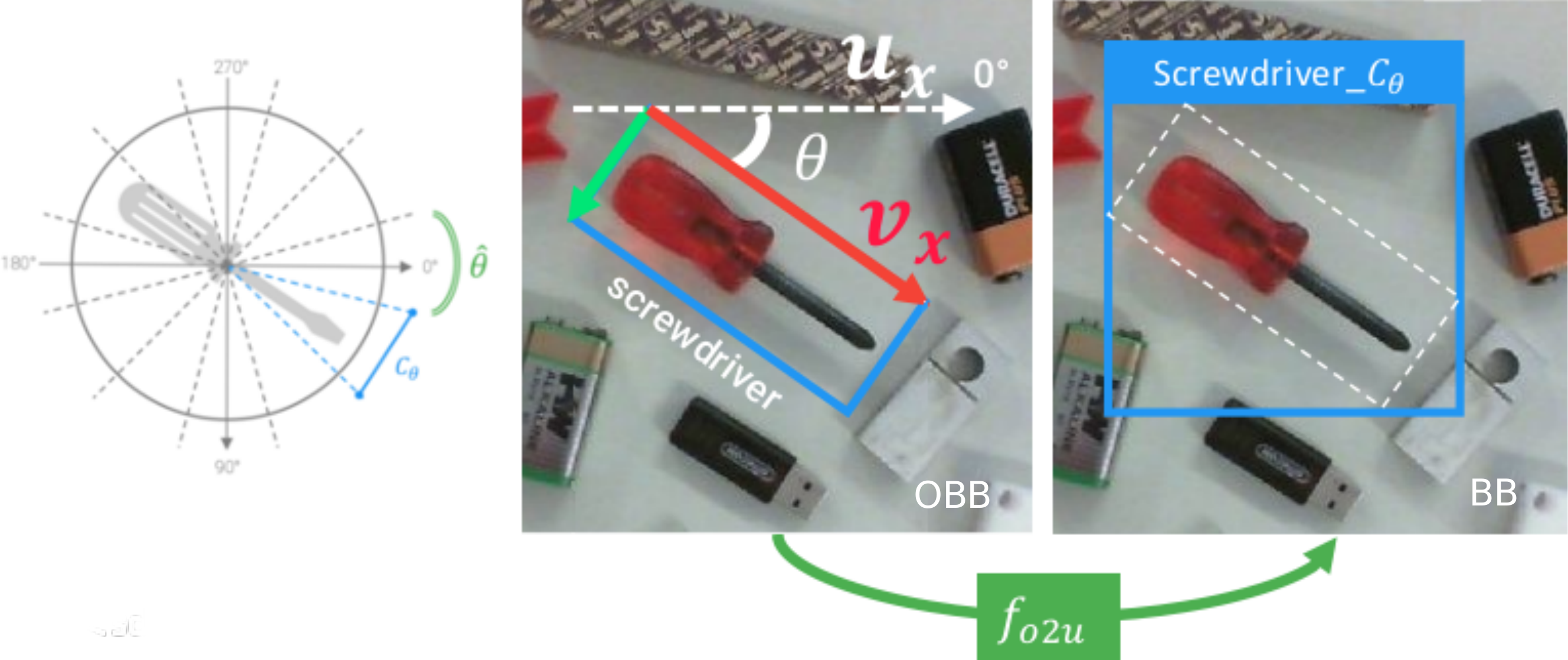}
  \caption{Graphic representation of the conversion between \emph{Oriented-to-Unoriented} bounding boxes. The real angle $\theta$ is converted in the index $C_{\theta}$ of the corresponding quantized bin among the $\nAngles = \ceil{ \frac{2\pi}{\hat{\theta}}}$ possible bins. }\label{fig:discretization}
   
\vspace*{-3mm}\end{figure}

\begin{figure}[]
  \centering
  \includegraphics[width=0.49\textwidth]{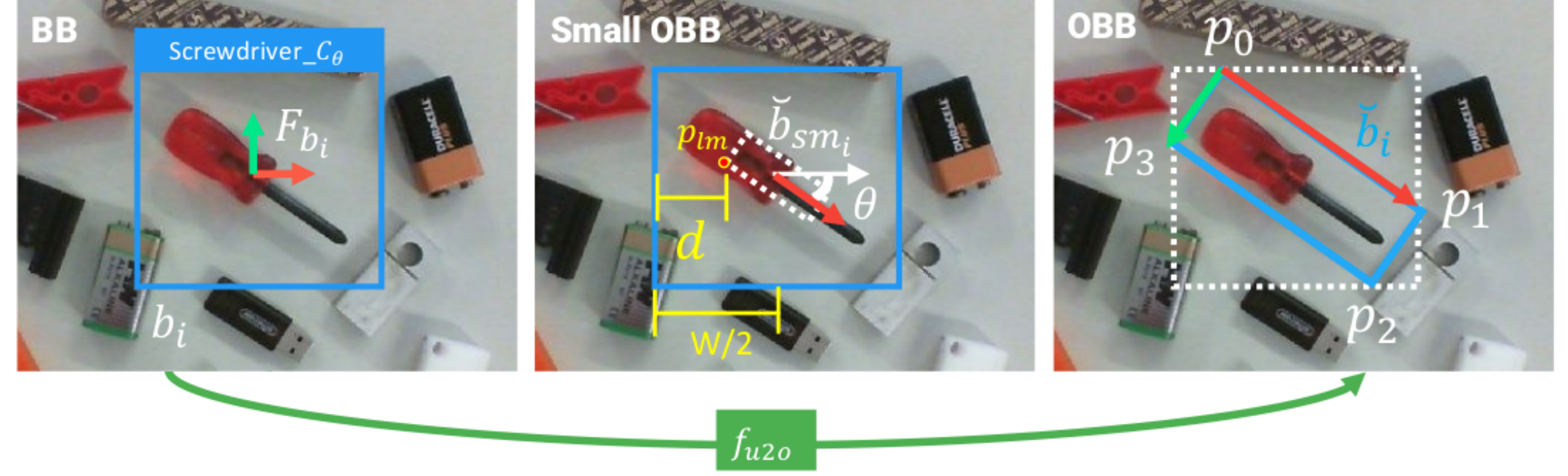}
  \caption{A graphical representation of the conversion between \emph{Unoriented-to-Oriented} bounding boxes. It is important to know the \emph{ratio} of the sought object in order to produce a commesurate oriented bounding box. }\label{fig:u2o}
\vspace*{-3mm}\end{figure}

As illustrated in \autoref{fig:teaser}, given an RGB image, the \algoname{} framework can produce a set of predictions $\breve{y}_i=\{ \breve{b}_i ,\theta_i, c_i \}$, where $\breve{b}_i  = \{x,y,w,h \} \in \Real^4$ represents the coordinates of the  Oriented Bounding Box (OBB in short) clockwise-rotated by an angle of $\theta_i$ and $c_i \in \IntegersP$ is the object class. As already mentioned, we leverage on a classical object detector, which outputs a set of simpler predictions $y_i=\{ b_i , \hat{c}_i \}$ where $b_i  = \{x,y,W,H \} \in \Real^4$ represents the coordinates of the \emph{unoriented} Bounding Box (BB in short) and $\hat{c}_i \in \IntegersP$ encodes, with our formulation, both the object class as well as orientation information. In \autoref{sec:function_o2u} we will explain how to transform an oriented prediction $\breve{y}_i$ into an un-oriented one, $y_i$, while in  \autoref{sec:u2o_function} we will describe the inverse procedure. Finally, in \autoref{sec:dataset_generation} we will explain how to generate Oriented Bounding Box labels for an entire video sequence by labeling just the first frame.

\subsection{$f_{o2u}$: The Oriented-to-Unoriented Function}\label{sec:function_o2u}

As stated before, our approach is an extension of a classical 2D Object Detector to make it capable of estimating also the orientation of a target object. We formulate the angle estimation problem as a classification task by simply quantizing the angular range  into $\nAngles$ bins and by expanding all the $C$ categories, managed by the object detector, into $C^\prime = \nAngles C$ new classes. 
Thus, as shown in \autoref{fig:discretization}, for each object instance, we can compute its real-valued angle with ${\theta = \arccos( \mathbf{u}_x \cdot  \mathbf{v}_x)}$, as the angle between the unit vector $\mathbf{v}_x$, directed as the first edge of the corresponding OOB, and the x axis of the image $\mathbf{u}_x$. 
Obviously the theta angle so defined, only for illustrative intent, is limited to the range $[0,\pi]$, for this reason it is necessary to calculate it using  

\begin{equation}
\theta = \begin{cases} 
\text{atan2}(v_{x_y},v_{x_x}), & \mbox{if } v_{x_y} >= 0 \\ 
2\pi + \text{atan2}(v_{x_y},v_{x_x}), & \mbox{if } v_{x_y} < 0
\end{cases}
\end{equation}

\noindent where ${(v_{x_y},v_{x_x})}$ are the components of the $\mathbf{v}_x$ vector.
The real-valued angle $\theta$ is then converted into the corresponding bin index $C_{\theta}$, with $C_{\theta} = \nint{\frac{\theta}{\hat{\theta}}} \in \{ 0, ... ,k \}$, where $\hat{\theta}$ is \emph{quantization step} so that $\nAngles = \ceil{\frac{2\pi}{\hat{\theta}}}$. In order to build an unique formulation to obtain the final converted class we can write:

\begin{equation}\label{eq:o2u_function}
f_{o2u}(c_i, \theta,\hat{\theta}) =  c_i k + C_{\theta} = c_i k + \nint{ \theta/\hat{\theta}} = \hat{c}_i
\end{equation}

\noindent where $f_{o2u}$ (\ie \emph{o2u}: \emph{Oriented-to-Unoriented}) is the function used to convert the original object class $c_i$ in the expanded class $\hat{c}_i$ which encodes not only the object type  but also its quantized orientation. The corresponding BB is computed simply by applying the minimum bounding box algorithm to the $4$ vertices of the original OOB.

\subsection{$f_{u2o}$: The Unoriented-to-Oriented Function}\label{sec:u2o_function}

Assuming $y_i=\{ b_i, \hat{c}_i \}$ the generic prediction of the Object Detector, where $\hat{c_i}$ is the predicted class (built with the \autoref{eq:o2u_function})  and $b_i$ the corresponding un-oriented bounding box, the purpose of the $f_{u2o}$ function, as depicted in \autoref{fig:teaser}, is to produce an equivalent oriented prediction $\breve{y}_i=\{ \breve{b}_i ,\theta_i, c_i \}$ where $c_i$ is the original class, mapping the object of belonging only, and $\breve{b}_i$ is an oriented bounding box (when omitted in images, the angle $\theta_i$ is represented, for simplification, with the red arrow oriented as the longest axis of an OBB). This procedure can be thought as the inverse of that described in \autoref{sec:function_o2u}. The function to determine the original class, $c_i$,  and the predicted angle $\theta_i$ is pretty simple:

\begin{equation}\label{eq:u2o_function}
f_{u2o}(\hat{c}_i, \hat{\theta})  = \begin{cases}
c_i = \floor{ \frac{\hat{c}_i}{k}}
 \\ 
 \theta_i = \hat{\theta}_i \cdot (\hat{c} \bmod k)
 \end{cases}
 \end{equation}

\noindent where $ \hat{\theta}$ is the same discretization step as used in the $f_{o2u}$ counterpart. Conversely, the estimation of the OBB ($\breve{b}_i$) given the simple BB ($b_i$) and the corresponding angle $\theta_i$ is not a trivial problem because of the infinite number of solutions with no constraints. However, for each object in the dataset we can compute the \emph{ratio} $r$ of its bounding box in a nominal condition (\eg when it exhibit $0^{\circ}$ in an image) and use it as a constraint to reduce the complexity of the procedure. If we define with $P_{\breve{b}_i}  = \{\mathbf{p}_0, \mathbf{p}_1, \mathbf{p}_2, \mathbf{p}_3 \} $ the set of OBB's $4$ expressed in the reference frame $F_{b_i}$ centered in the bounding box $b_i$ (the clockwise order, as depicted in the last frame of \autoref{fig:u2o}, has to be reliable in order to have that $\mathbf{v}_x = \frac{\mathbf{p}_1 -  \mathbf{p}_0}{\rvert \rvert \mathbf{p}_1 -  \mathbf{p}_0 \lvert \lvert}$), the ratio can be computed easily as $r = \frac{\mathbf{p}_1 -  \mathbf{p}_0}{\mathbf{p}_3 -  \mathbf{p}_0}$. Thanks to the object aspect ratio we can execute the pipeline depicted in \autoref{fig:u2o} to build an OBB: starting from the original BB and an angle $\theta$, we superimpose a small version of the corresponding OBB, $\breve{b}_{sm_i}$ (built only using the aspect ratio information), in the center of the BB rotating it by the provided angle; we estimate as ${d = \lvert \frac{W}{2} + p_{lm_x} \rvert}$ the distance between the leftmost vertex of the OBB, $\mathbf{p}_{lm}$, and the left edge of the original BB; we enlarge $\breve{b}_{sm_i}$ by a scaling factor $s$, in order to have $d=0$.
Therefore, taking the example of \autoref{fig:u2o}, if $\lmpoint \coloneqq \mathbf{p}_3 $, we can simplify the above mentioned distance condition, obtaining the new desired $x$ coordinate of $\mathbf{p}_3$ as  ${\hat{p}_{3_x}= - \frac{W}{2}}$. And then reformulating it as a scaling problem $\hat{p}_{3_x} = s \cdot p_{3_x}$ \noindent we get that $s = -\frac{W}{2p_{3_x}}$. The factor $s$ can be used to generate a  scaling matrix and transform all 4 vertices consistently. 

As mentioned above, due to the error introduced by the Object Detector in the estimation of $b_i$ instances, the construction of a BB from an OBB is not perfectly invertible, so the algorithm just proposed tries to minimize one of the many possible constraints (the proximity of the \emph{left-most} point). Surely an optimization algorithm could take into account more than one constraint (e.g. the proximity of all $4$ vertices) but in our case this approach is sufficiently precise and fast to test the rest of the pipeline. 

\subsection{Automatic Dataset Generation}\label{sec:dataset_generation}

In \autoref{sec:introduction} we underlined that it is important to provide a smart solution to collect training data for data-driven models to be deployed in real industrial applications. For this reason, we endowed \algoname{} with an automated labeling tool based on  video sequences. The hypotheses allowing the tool to work well are: 

\begin{enumerate}
\item  the video sequence frames a tabletop scene with the image plane as parallel as possible to the supporting surface;
\item  camera movements have to be, as much as possible, only rotational (\emph{Rotate} around the camera z axis) and translational (\emph{Lift} along the z axis);
\item the height of target objects has to be mostly uniform.
\end{enumerate}

\begin{figure}[t]
  \centering
  \includegraphics[width=0.7\columnwidth]{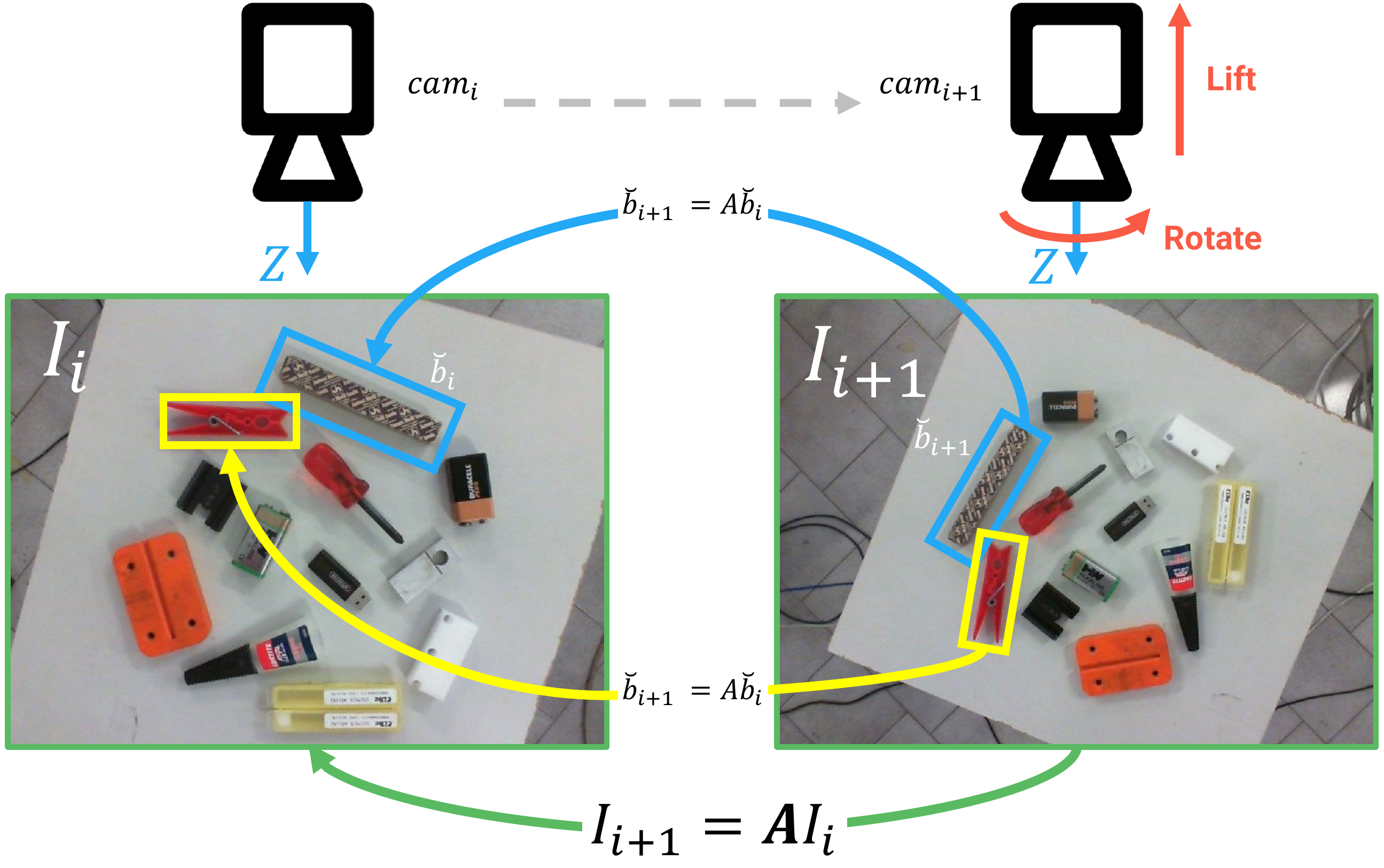}
  \caption{For each pair of consecutive frames $I_i,I_{i+1}$ it is possible to estimate the rigid transformation $\mathbf{A}$ such that $I_{i+1}=\mathbf{A}I_{i}$, by means of a 2D feature matching pipeline, like \eg ORB\cite{rublee2011orb}. The same transformation $\mathbf{A}$ can be applied to each OOB label. }\label{fig:labeling}
\vspace*{-3mm}\end{figure}

\noindent \autoref{fig:labeling} exemplifies the above requirements by showing two consecutive frames of a suitable video sequence. The figure shows  clearly how restricted camera movements (\ie lift and rotate) leads to a controlled rigid transformation $\mathbf{A}$ between the two consecutive images $I_i, I_{i+1}$ such that $I_{i+1} = \mathbf{A}I_i$. The same rigid transformation $A$ can be applied to each OOB $\breve{b}_i$ present in the image $I_i$ so as to obtain a new set of OOB such that $\breve{b}_{i+1} = \mathbf{A} \breve{b}_i$. This procedure can be repeated for each consecutive pair of images in the video sequence, it is therefore clear how the sole human intervention is to create the OOB labels in the first frame $I_0$. To estimate the rigid transformation $\mathbf{A}$ any 2D feature-based matching pipeline, described in \autoref{sec:related}, may be used. We decided to use ORB \cite{rublee2011orb}, a patent-free solution, to make our software completely open-source$^1$.


\section{RESULTS }
\label{sec:results}

In this section we will describe both data and models used in our experiments in order to maximize reproducibility. In \autoref{sec:experimental_dataset} we will introduce a novel dataset, dubbed the \emph{\datasetName{}}, used in our quantitative and qualitative experiments. In \autoref{sec:object_detector} we will describe how the core object detector has been trained over the \datasetName{} in order to obtain different declensions of the more complex \algoname{} Detector. To prove the effectiveness of our solution, in \autoref{sec:performances} we analyze the absolute \algoname{}'s performances, while in \autoref{sec:veterans_vs} we present a detailed comparison between \algoname{} and the state-of-the-art approaches based on handcrafted features. Finally, in \autoref{sec:qualitative}, a qualitative evaluation of our approach within a typical robotic task is provided.

\subsection{Experimental setup: \datasetName}\label{sec:experimental_dataset}

\begin{figure*}[]
  \centering
  \includegraphics[width=0.6\textwidth]{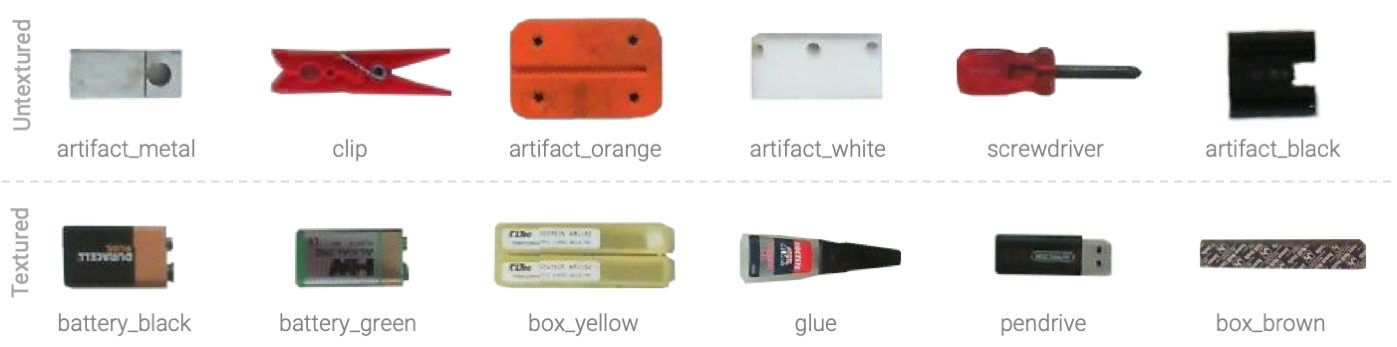}
  \caption{\datasetName{} objects grouped in \emph{Textured} and \emph{Untextured}.}\label{fig:dataset}
  
\vspace*{-3mm}\end{figure*}


\begin{figure}[]
  \centering
  \includegraphics[width=0.9\columnwidth]{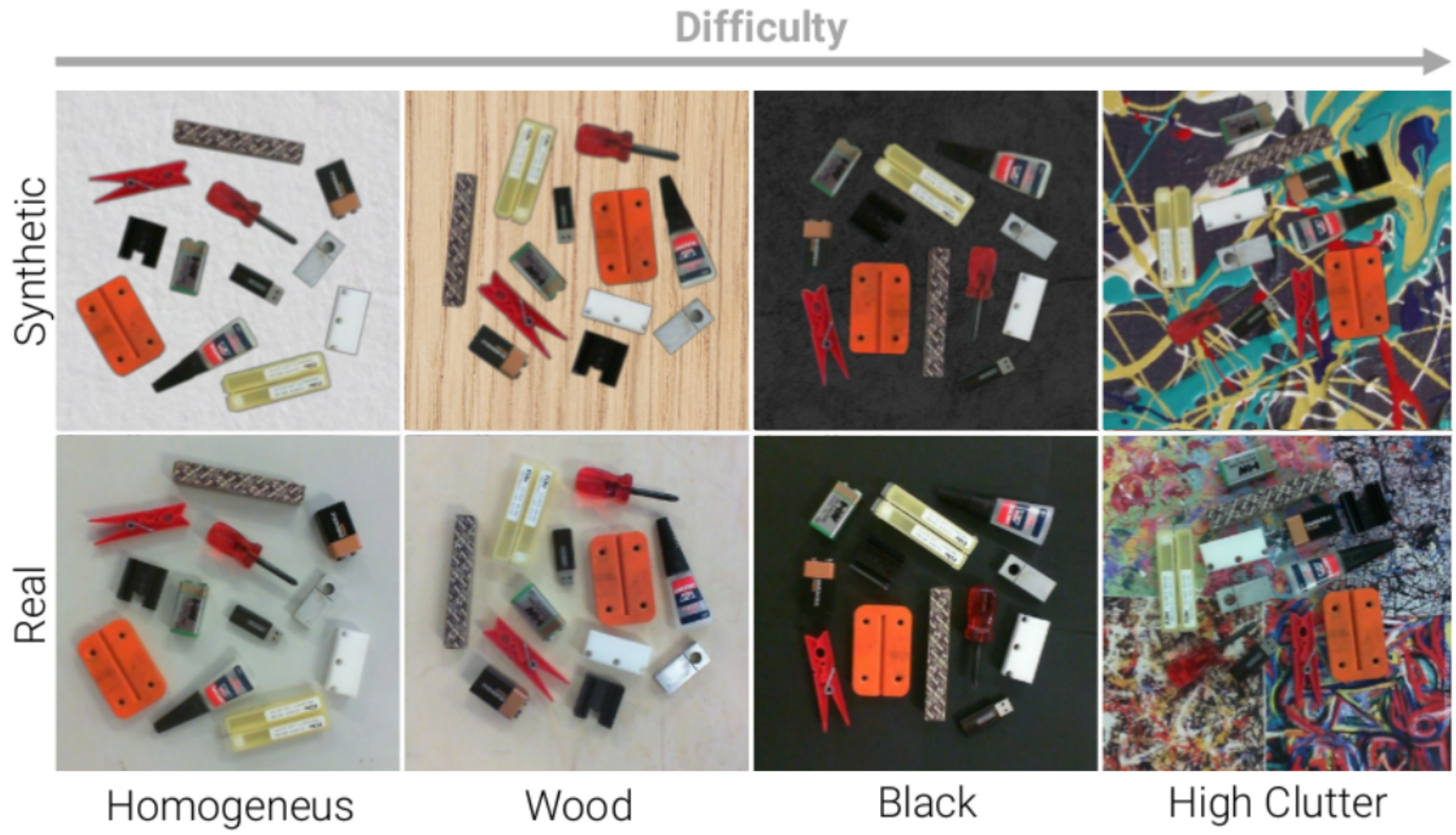}
  \caption{In the left column: $4$ samples coming from the \datasetName{} one for each background category. The right column shows the synthetic version of each sample. }\label{fig:dataset_samples}
\vspace*{-6mm}\end{figure} 

Thanks to the technique examined in \autoref{sec:dataset_generation}, we created our an experimental --public-- dataset with $12$ objects equally divided into \emph{Textured} and \emph{Untextured} (as shown in \autoref{fig:dataset}), submerged in several challenging scenes, in order to extensively compare our approach with classical ones.
We collected $15$ tabletop scenes, with randomly arranged objects, featuring different backgrounds: $3$ scenes with \emph{homogeneous} background; $3$ scenes with \emph{wood}; $3$ scenes with \emph{black} background and $5$ scenes with an high-clutter background (several prints of \emph{Pollock}'s painting). This increasing variability is used in order to achieve \emph{Domain Randomization} \cite{tobin2017domain} during training and very challenging scenes during the test. 
These $4$ different scene categories are shown in \autoref{fig:dataset_samples}, where they are intentionally presented in increasing order of complexity. We collected a total of $7155$ self-labeled images (strictly speaking, we manually labeled only $15$ images, one per scene).
Moreover, by using the \emph{ground truth} coming from the whole dataset we also built a \emph{Synthetic} version of it just by stitching one version of each model with the same arrangement proposed by the original dataset (right column in \autoref{fig:dataset_samples} illustrates synthetic version of the real images in the left column). This version of the dataset is built in order to reproduce a version of the training data in which the distribution of the objects in terms of position and orientation is identical to the real one (also the backgrounds are intentionally similarly synthetized) but the variance of objects semblance, in the images space, is purposely kept low: \ie stitching the same version of the object synthetically rotated is, under our hypothesis, less effective than produce real rotated viewpoints, especially dealing with deep neural models, not to mention that even the variations of light conditions and perspective are not correctly captured by a synthetic dataset generated this way.

\subsection{Deep Object Detector}\label{sec:object_detector}

As reference for our benchmarks we used YOLOv3 \cite{redmon2018yolov3}, a state-of-the-art Object Detector based on CNNs. We fine-tuned the YOLOv3 model, pretrained on ImageNet \cite{krizhevsky2012imagenet}, with the \algoname{} approach using $13$ scenes of the \datasetName{} (about 6200 labeled images) with a 80\%/20\% split for training and test. The training strategy is to freeze weights of the network's feature extractor (namely, the backend), for $2$ epochs training only the object detection layers (namely, the frontend), then fine-tuning the whole architecture for $50$ further epochs, with a learning rate of $0.001$. The two remaining scenes of the dataset (\ie one with wood background, thought as a simple testbench, and one with high clutter background thought as a complex one), never used during the training phase are used to evaluate the performance of the whole system. We will call these two scenes \emph{Simple Scene} (477 images) and \emph{Hard Scene} (477 images) in the rest of this section. 

We trained several models using different $\hat{\theta}$ (\ie angle discretization parameter, as described in \autoref{sec:function_o2u}), thereby producing different declensions of the detector useful in understanding how the discretization factor goes to affect performances. For the sake of convenience, we will adopt the short nickname $\algoname{}_{\alpha}$ to identify a \algoname{} model trained using the discretization angle $\hat{\theta}=\alpha$ (\eg  $\algoname{}_{10}$ identifies a model trained by quantizing the whole \emph{angle turn} in $36$ bins of $10^{\circ}$ each). Moreover, we add the letter $S$ to the above mentioned notation (\eg $\algoname{}_{\alpha}^S$) to represent the same model trained with the synthetic version of the \datasetName{}.

\subsection{\algoname{} performances}\label{sec:performances}


\begin{figure*}
   \def \colsize {0.35}
	\centering
	\begin{tabular}{cc}
		\includegraphics[width=\colsize\textwidth]{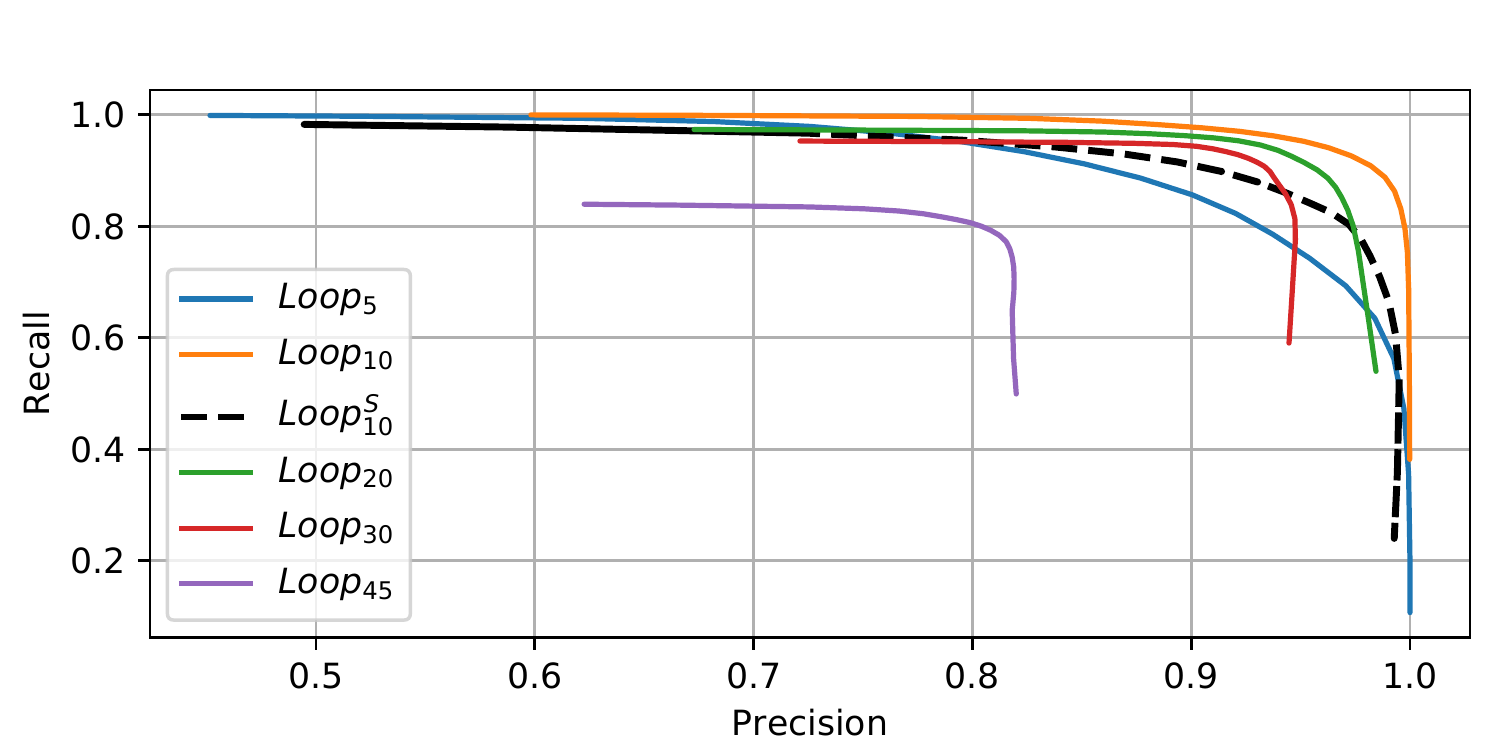} &
		\includegraphics[width=\colsize\textwidth]{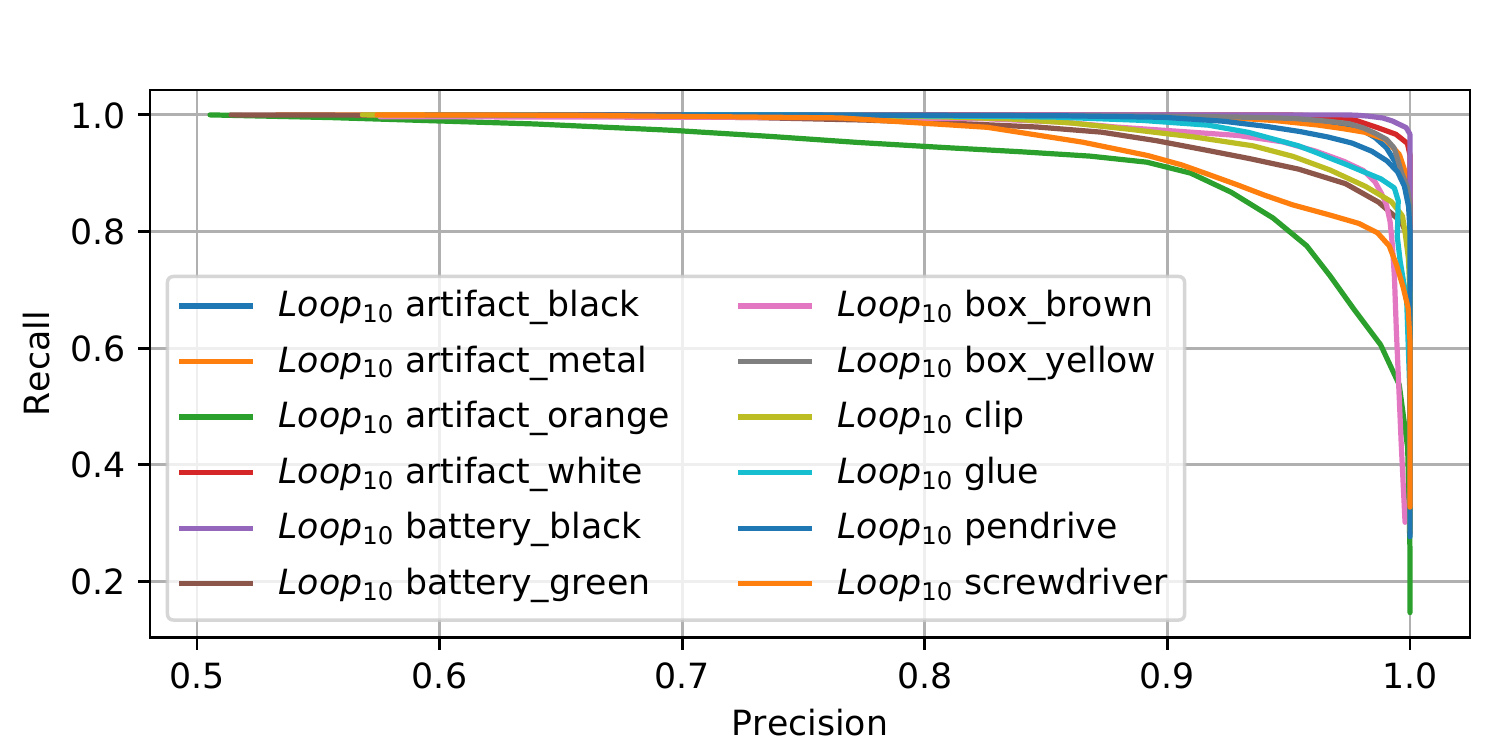} \\
		\textbf{(a)} Simple Scene & \textbf{(d)} Simple Scene \\
		
		\includegraphics[width=\colsize\textwidth]{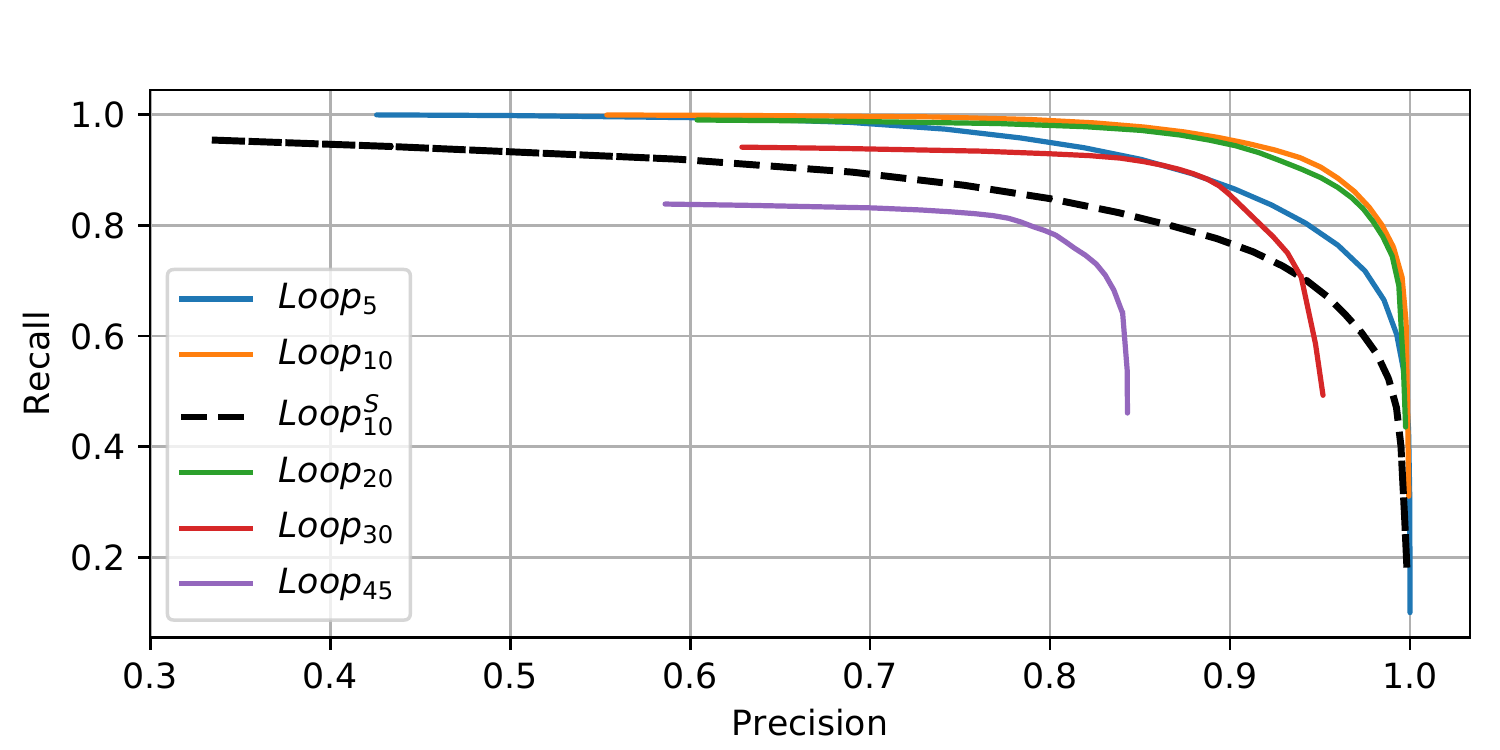} &
		\includegraphics[width=\colsize\textwidth]{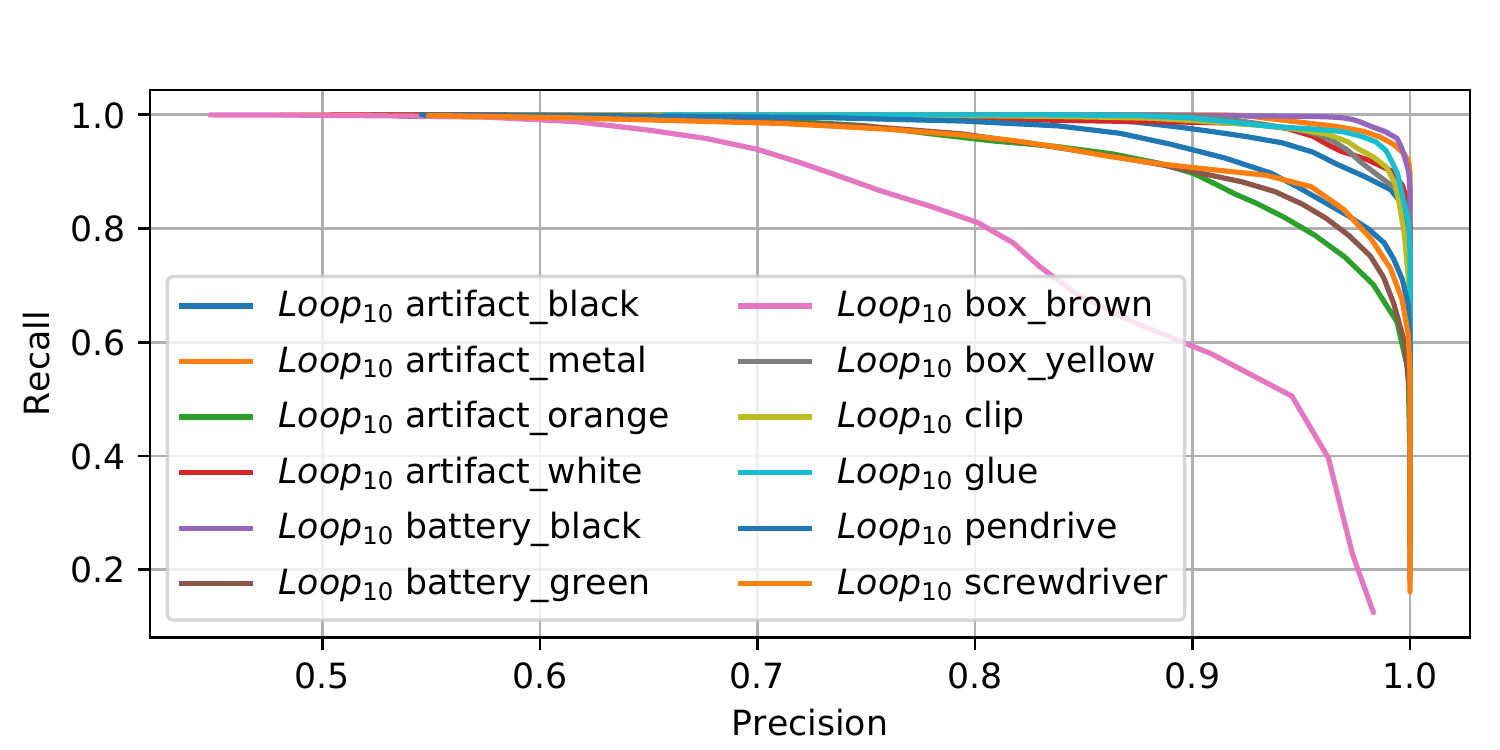} \\
		\textbf{(b)} Hard Scene & \textbf{(e)} Hard Scene \\
		
		\includegraphics[width=\colsize\textwidth]{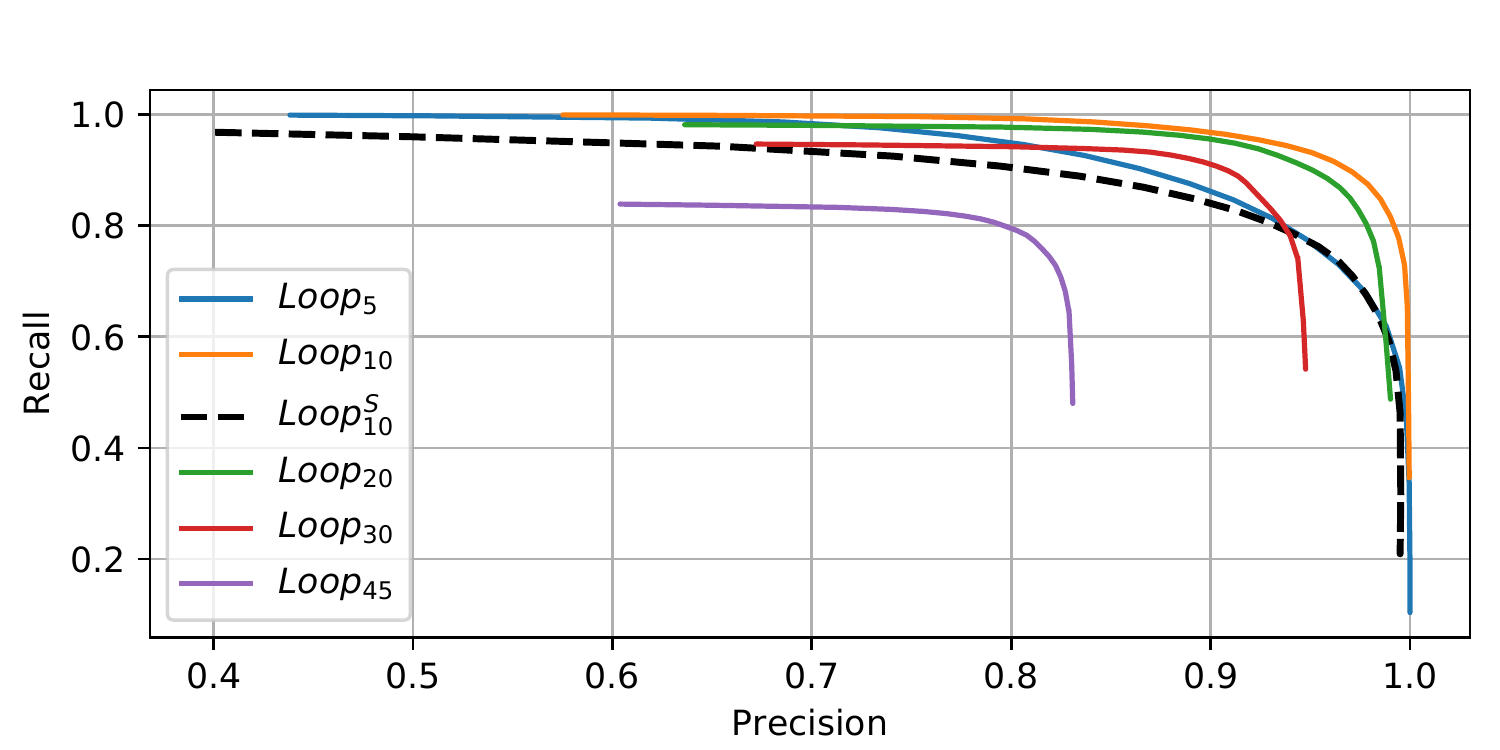}  & 
		\includegraphics[width=\colsize\textwidth]{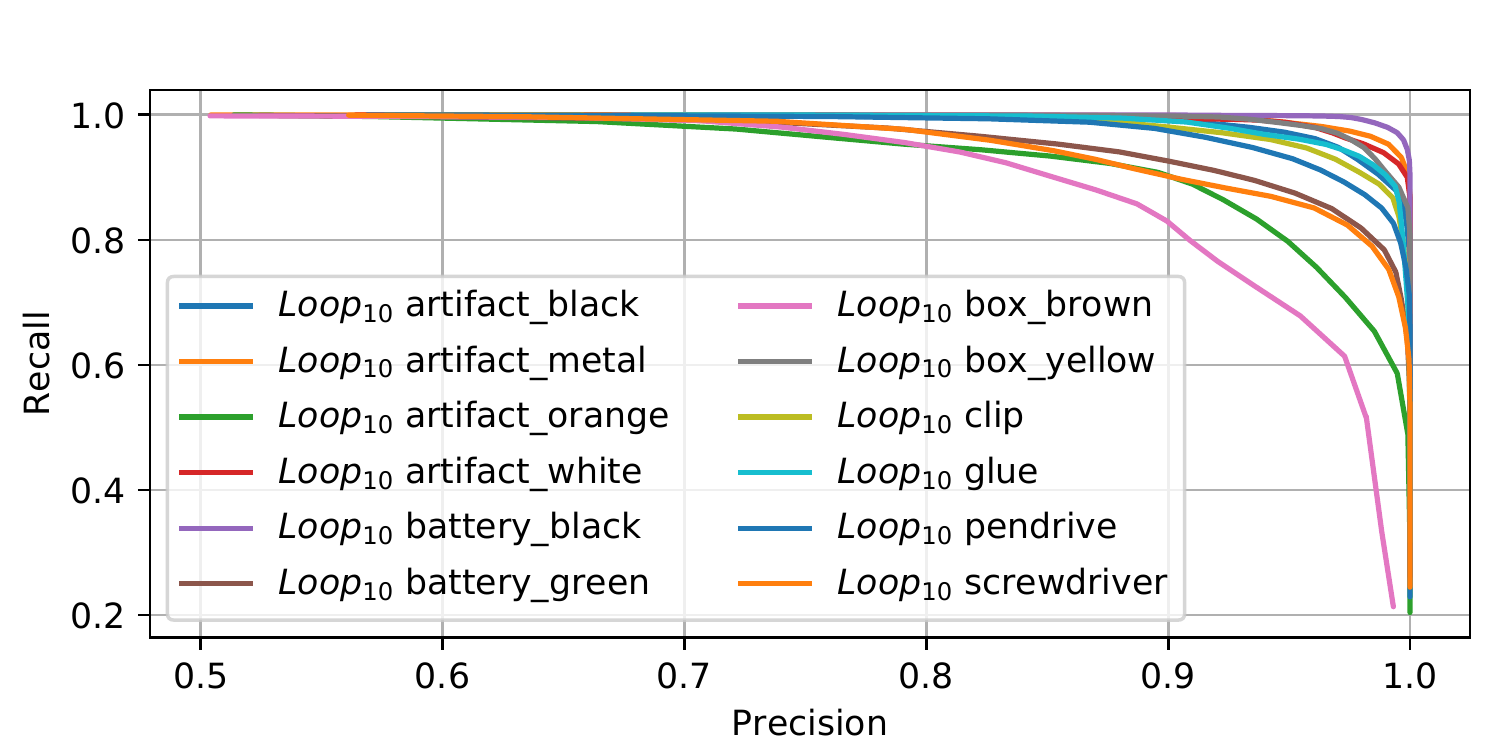} \\
		\textbf{(c)} Overall & \textbf{(f)} Overall
	\end{tabular}
	\caption{Precision/Recall curves for the \algoname{} approach. These plots have been produces by varying the threshold of the predictions \emph{confidence} between $0.0$ and $1.0$ with a step of $0.05$. The first column compares various discretization angles. The second column shows performances of the best version, $\algoname{}_{10}$, for each single object in the dataset. }
	\label{fig:perfomances_plots} 
\vspace*{-3mm}\end{figure*}

In the claim of this work we said that with the \algoname{} approach it is possible to build effortlessly an Object Detector, based on CNNs suitable to real industrial applications. In this section, indeed, we evaluate \algoname{} over the \datasetName{} to understand if its performances are good enough to embody it into a reliable industrial system.

\autoref{fig:perfomances_plots} groups together several Precision/Recall curves obtained by varying the threshold over the confidence output of the YOLOv3 model.  \autoref{fig:perfomances_plots} (a), (b) and (c) depict the precision/recall of several detectors (\ie each of which trained with a different angle discretization $\hat{\theta}$) over the \emph{Simple Scene}, \emph{Hard Scene} and both, respectively. In the same plots also the $\algoname{}_{10}^S$ model, thus trained only over synthetic data, is depicted in order to assess its performance compared to its counterparty trained on real data (\ie $\algoname{}_{10}$). \autoref{tab:loop_map} resumes the \emph{mean average precision} for the considered models: in this concise summary it is even clearer how deploying a  synthetic dataset is significantly less effective than leveraging on real imagery, especially when it comes to very complex real situations.  

From these results we can conclude that $\algoname{}_{10}$ is the best version of our detector. The $\algoname{}_{5}$  and $\algoname{}_{20}$ versions resulted slightly worse. This analysis shows, clearly, how the angle discretization parameter $\hat{\theta}$ affects performances in both direction: too small a value may cause a dramatic increase of detector categories $C^\prime$ (\eg with $\hat{\theta}=5$ we have ${C^\prime = \frac{360}{5}12 = 864}$); on the other hand, if $\hat{\theta}$ is too high, the angle prediction is subject to a large discretization error producing low IoU scores. Moreover, training on synthetic data only ($\algoname{}_{10}^S$), although acceptable, is worse than training on real data. With regard to single objects precision/recall, as will be described in more detail later, symmetric objects are -- unsurprisingly -- somewhat confusing the algorithm.

\begin{table}[]
	\centering
	\resizebox{0.5\columnwidth}{!}{

	\begin{tabular}{c|c|c|c}
\toprule
Model & Simple & Hard & Overall \\
\midrule
$\algoname{}_{5}$ 		& 0.96 & 0.96 & 0.96 \\
$\algoname{}_{10}$ 		& \textbf{0.99} & \textbf{0.98} & \textbf{0.99} \\
$\algoname{}_{10}^S$ 	& 0.95 & 0.86 & 0.92 \\
$\algoname{}_{20}$ 		& 0.95 & 0.97 & 0.96 \\
$\algoname{}_{30}$ 		& 0.90 & 0.88 & 0.89 \\
$\algoname{}_{45}$ 		& 0.69 & 0.70 & 0.69 \\

\bottomrule 
\end{tabular}

	} 
	\vspace{0.2cm}
	\caption{The mean average precision (\emph{mAP}) computed for each model, across all objects, for the \emph{Simple Scene}, \emph{Hard Scene} and both.}
	\label{tab:loop_map}
	
\vspace*{-7mm}\end{table}

\subsection{Hand-crafted features  vs Deep Learning}\label{sec:veterans_vs}

We used our \datasetName{} to test the \algoname{} approach compared with the state-of-the-art algorithms based on hand-crafted features. The first competitor is, certainly, SIFT \cite{lowe1999object}, one of the most used feature-based approach for \emph{Textured} objects detection. For the \emph{Untextured} counterpart, we chose BOLD \cite{tombari2013bold}, which uses highly repeatable geometric primitives as 2D features in order to perform object detection also with monochrome targets. Both methods are implemented in the same object detection pipeline as described in \cite{tombari2013bold}: 1) key-points detection; 2) key-point description; 3) features correspondences validated through Generalized Hough Transform (as described in \cite{lowe1999object})and 4) Pose Estimation.
\noindent The Pose Estimation in this case is modified in order to obtain a Rigid Transform instead of a Full Affine (homography). In this way we are sure that both detectors will yield predictions similar to  $\breve{y}_i=\{ \breve{b}_i ,\theta_i, c_i \}$ without producing distorted bounding boxes, as it is the case of homographies.

We compare the output of the previous two pipelines with the output directly obtained with the \algoname{} approach. We measured the performances with the classical precision/recall metric computed taking into account the \emph{Intersection Over Union} (IoU) of the predicted OBB: each detection, of a generic algorithm, is counted as \emph{True Positive} if its IoU is $>0.5$ compared with the corresponding ground truth OBB. With this metric we can analyse: Precision, Recall, FScore and Average IoU for each algorithm. We introduce also an additional term called \emph{Oriented Intersection Over Union} (OIoU) which measures the classical IoU commesurate to the effectiveness of the algorithm in predicting the correct angle. Formally $
\text{OIoU} = \text{IoU} \times \text{max}\left(\frac{\mathbf{v}_x \cdot \mathbf{\hat{v}}_x}{ \| \mathbf{v}_x \|\|  \mathbf{\hat{v}}_x \|},0\right)$,  so the Oriented IoU is inversely proportional to the angle between the two $\mathbf{v}_x$ of the ground truth and the predicted OBB; it is forced to be between $0$ and $1$, with the $max$ operator, in order to discard a priori opposed angles. The OIoU term is introduced to measure the capability of each algorithm in distinguishing quite symmetric objects: in the \datasetName{}, for instance, the two objects \emph{artifact\_orange} and  \emph{box\_brown} seems very symmetrical even thought they are not. In such cases, if the angle prediction is completely wrong, the IoU may be high but the OIoU is very low.

\begin{figure}[t]
  \centering
  \includegraphics[width=0.45\textwidth]{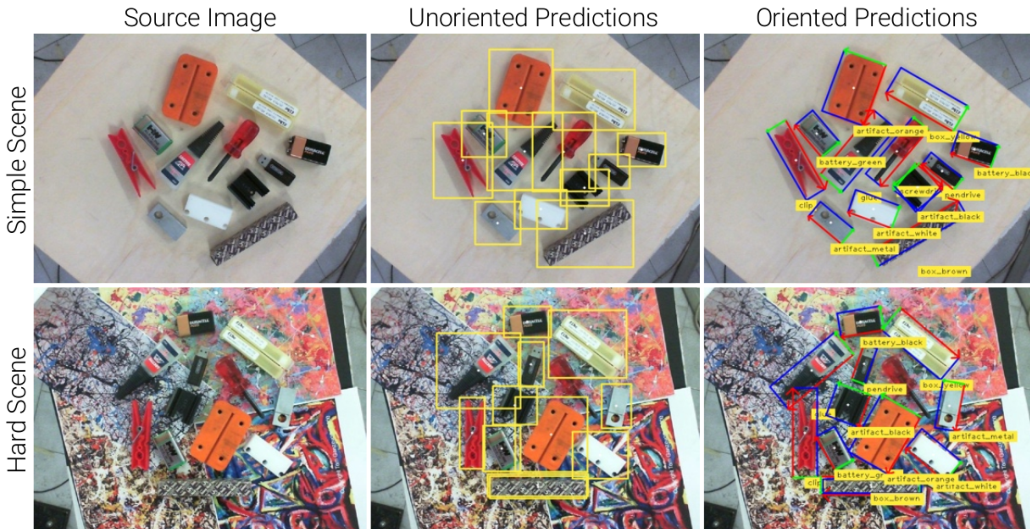}
  \caption{This picture shows  the real output of our software for the simple and hard scene, respectively. The middle column shows Unoriented Predictions (BB) coming from the generic object detector, while the right column shows the derived Oriented Predictions (OBB). }\label{fig:qualitative}
 
\vspace*{-3mm}\end{figure}

Under the assumption that, as deduced in \autoref{sec:performances}, the model with the best trade-off between precision and recall is $\algoname{}_{10}$, we used the latter to compete with the state-of-the-art. In \autoref{tab:results_simple} and \autoref{tab:results_hard} several comprehensive results are shown. The first table compares SIFT and BOLD with $\algoname{}_{10}$ and $\algoname{}_{10}^S$ dealing with the \emph{Simple Scene}, \ie the scene with a low clutter background. The second table, instead, deals with the \emph{Hard Scene} containing an highly cluttered background. Both tables contain performances dealing with  each object, a summary for Textured and Untextured objects and an Overall index. As vouched by these experimental  results, \algoname{} outperforms both SIFT and BOLD in both scenarios: our approach is able to cope with both Textured and Untextured objects seamlessly and is very robust to the high level of clutter in the \emph{Hard Scene}. On the other hand, as pointed out in \autoref{tab:results_simple}, the OIoU index shows a slight fall of \algoname{} when dealing with symmetric objects (\eg the \emph{box\_brown} OIoU is $0$). This conceptual problem can be easily resolved treating symmetrical objects differently during the angle discretization process, described in \autoref{sec:function_o2u}, with a formulation similar to the one introduced in \cite{sundermeyer2018implicit}. A qualitative evaluation is present in \autoref{fig:qualitative} featuring a real output of the \algoname{} detector on two samples randomly picked between the Simple and the Hard scenes subsets.

\begin{table*}[]
	\centering
	\resizebox{\textwidth}{!}{%

	\begin{tabular}{l|cccc|cccc|cccc|cccc|cccc}
\toprule
Index & \multicolumn{4}{l|}{Precision} & \multicolumn{4}{l|}{Recall} & \multicolumn{4}{l|}{FScore} & \multicolumn{4}{l|}{IOU} & \multicolumn{4}{l}{OIOU} \\
\hline
Method &      Sift &  Bold &  $\algoname{}_{10}$ & $\algoname{}_{10}^S$ &   Sift &  Bold &  $\algoname{}_{10}$ & $\algoname{}_{10}^S$ &Sift &  Bold &  $\algoname{}_{10}$ & $\algoname{}_{10}^S$ &Sift &  Bold &  $\algoname{}_{10}$ & $\algoname{}_{10}^S$ &Sift &  Bold &  $\algoname{}_{10}$ & $\algoname{}_{10}^S$   \\
\midrule

artifact\_black & 0.50 & 0.84 & \textbf{0.99} & 0.97 & 0.00 & 0.81 & 0.96 & \textbf{0.98} & 0.00 & 0.82 & \textbf{0.97} & 0.97 & 0.75 & 0.77 & \textbf{0.85} & 0.84 & 0.75 & 0.75 & \textbf{0.85} & 0.84 \\
artifact\_metal & 0.00 & 0.00 & \textbf{0.97} & 0.95 & 0.00 & 0.00 & \textbf{0.98} & 0.86 & 0.00 & 0.00 & \textbf{0.97} & 0.90 & 0.00 & 0.00 & \textbf{0.77} & 0.77 & 0.00 & 0.00 & \textbf{0.77} & 0.77 \\
artifact\_orange & 0.54 & \textbf{0.96} & 0.91 & 0.78 & 0.12 & \textbf{0.96} & 0.91 & 0.88 & 0.20 & \textbf{0.96} & 0.91 & 0.83 & 0.71 & \textbf{0.86} & 0.86 & 0.86 & 0.62 & 0.43 & \textbf{0.78} & 0.37 \\
artifact\_white & 0.62 & 0.00 & \textbf{0.97} & 0.90 & 0.85 & 0.00 & \textbf{1.00} & 0.98 & 0.72 & 0.00 & \textbf{0.98} & 0.94 & 0.78 & 0.00 & \textbf{0.84} & 0.80 & 0.78 & 0.00 & \textbf{0.84} & 0.79 \\
clip & 0.78 & 0.60 & \textbf{0.95} & 0.91 & 0.63 & 0.60 & \textbf{0.93} & 0.92 & 0.70 & 0.60 & \textbf{0.94} & 0.91 & 0.76 & 0.76 & \textbf{0.81} & 0.75 & 0.76 & 0.74 & \textbf{0.81} & 0.74 \\
screwdriver & 0.76 & 0.58 & 0.95 & \textbf{0.97} & 0.05 & 0.58 & 0.85 & \textbf{0.89} & 0.10 & 0.58 & 0.89 & \textbf{0.93} & 0.71 & 0.73 & \textbf{0.81} & 0.79 & 0.71 & 0.72 & \textbf{0.81} & 0.79 \\
 \hline Untextured & 0.53 & 0.50 & \textbf{0.95} & 0.91 & 0.28 & 0.49 & \textbf{0.94} & 0.92 & 0.29 & 0.49 & \textbf{0.95} & 0.91 & 0.62 & 0.52 & \textbf{0.82} & 0.80 & 0.60 & 0.44 & \textbf{0.81} & 0.72 \\ \hline
battery\_black & 0.48 & 0.82 & \textbf{0.98} & 0.94 & 0.81 & 0.82 & 1.00 & \textbf{1.00} & 0.60 & 0.82 & \textbf{0.99} & 0.97 & 0.76 & 0.71 & \textbf{0.84} & 0.83 & 0.75 & 0.68 & \textbf{0.84} & 0.83 \\
battery\_green & 0.65 & 0.44 & \textbf{0.93} & 0.90 & 0.83 & 0.43 & \textbf{0.93} & 0.91 & 0.73 & 0.43 & \textbf{0.93} & 0.90 & 0.69 & 0.70 & 0.85 & \textbf{0.85} & 0.69 & 0.59 & \textbf{0.82} & 0.79 \\
box\_brown & 0.65 & 0.68 & \textbf{0.97} & 0.87 & 0.77 & 0.68 & 0.92 & \textbf{0.95} & 0.71 & 0.68 & \textbf{0.95} & 0.91 & \textbf{0.80} & 0.75 & 0.74 & 0.73 & \textbf{0.80} & 0.65 & 0.00 & 0.02 \\
box\_yellow & 0.49 & \textbf{1.00} & 0.98 & 0.98 & 1.00 & \textbf{1.00} & 0.98 & 0.59 & 0.66 & \textbf{1.00} & 0.98 & 0.73 & 0.82 & 0.81 & \textbf{0.86} & 0.80 & 0.82 & 0.81 & \textbf{0.86} & 0.79 \\
glue & 0.37 & \textbf{0.99} & 0.97 & 0.95 & \textbf{1.00} & 0.99 & 0.91 & 0.97 & 0.54 & \textbf{0.99} & 0.94 & 0.96 & 0.80 & 0.81 & 0.82 & \textbf{0.83} & 0.80 & 0.80 & 0.82 & \textbf{0.83} \\
pendrive & 0.78 & 0.91 & \textbf{0.98} & 0.92 & 0.36 & 0.91 & \textbf{0.94} & 0.85 & 0.49 & 0.91 & \textbf{0.96} & 0.89 & 0.75 & 0.74 & \textbf{0.76} & 0.73 & 0.74 & 0.73 & \textbf{0.76} & 0.71 \\
 \hline Textured & 0.58 & 0.79 & \textbf{0.97} & 0.92 & 0.75 & 0.78 & \textbf{0.94} & 0.86 & 0.61 & 0.78 & \textbf{0.95} & 0.88 & 0.77 & 0.76 & \textbf{0.81} & 0.79 & \textbf{0.77} & 0.71 & 0.67 & 0.64 \\ \hline
global & 0.54 & 0.72 & \textbf{0.96} & 0.91 & 0.54 & 0.65 & \textbf{0.94} & 0.90 & 0.54 & 0.68 & \textbf{0.95} & 0.91 & 0.77 & 0.77 & \textbf{0.82} & 0.80 & \textbf{0.77} & 0.70 & 0.75 & 0.69 \\

\bottomrule \end{tabular}

	}
	\vspace{0cm}
	\caption{Performances of our approach, compared to SIFT\cite{lowe1999object}  and BOLD\cite{tombari2013bold}, in the \emph{Simple Scene} scenario (boldface text highlights the best score in the related pane). The overall Precision/Recall index of \algoname{} is about 96\%/94\% showing its flexibility in general purpose real applications. }
	\label{tab:results_simple}
	
	\vspace{-3mm}
\end{table*}

\begin{table*}[]
	\centering
	\resizebox{\textwidth}{!}{%

	\begin{tabular}{l|cccc|cccc|cccc|cccc|cccc}
\toprule
Index & \multicolumn{4}{l|}{Precision} & \multicolumn{4}{l|}{Recall} & \multicolumn{4}{l|}{FScore} & \multicolumn{4}{l|}{IOU} & \multicolumn{4}{l}{OIOU} \\
\hline
Method &      Sift &  Bold &  $\algoname{}_{10}$ & $\algoname{}_{10}^S$ &   Sift &  Bold &  $\algoname{}_{10}$ & $\algoname{}_{10}^S$ &Sift &  Bold &  $\algoname{}_{10}$ & $\algoname{}_{10}^S$ &Sift &  Bold &  $\algoname{}_{10}$ & $\algoname{}_{10}^S$ &Sift &  Bold &  $\algoname{}_{10}$ & $\algoname{}_{10}^S$   \\
\midrule

artifact\_black & 0.00 & 0.01 & \textbf{0.96} & 0.90 & 0.00 & 0.01 & \textbf{0.94} & 0.90 & 0.00 & 0.01 & \textbf{0.95} & 0.90 & 0.00 & 0.64 & \textbf{0.80} & 0.75 & 0.00 & 0.05 & \textbf{0.80} & 0.74 \\
artifact\_metal & 0.43 & 0.00 & \textbf{0.97} & 0.83 & 0.01 & 0.00 & \textbf{0.98} & 0.93 & 0.02 & 0.00 & \textbf{0.97} & 0.87 & 0.70 & 0.00 & \textbf{0.82} & 0.76 & 0.70 & 0.00 & \textbf{0.82} & 0.76 \\
artifact\_orange & 0.75 & 0.00 & \textbf{0.91} & 0.78 & 0.41 & 0.00 & 0.88 & \textbf{0.92} & 0.53 & 0.00 & \textbf{0.89} & 0.85 & 0.85 & 0.00 & \textbf{0.87} & 0.86 & \textbf{0.84} & 0.00 & 0.82 & 0.68 \\
artifact\_white & 0.73 & 0.00 & \textbf{0.96} & 0.80 & 0.10 & 0.00 & 0.96 & \textbf{0.98} & 0.18 & 0.00 & \textbf{0.96} & 0.88 & 0.68 & 0.00 & \textbf{0.80} & 0.80 & 0.68 & 0.00 & \textbf{0.79} & 0.76 \\
clip & 0.00 & 0.40 & \textbf{0.96} & 0.91 & 0.00 & 0.40 & \textbf{0.97} & 0.87 & 0.00 & 0.40 & \textbf{0.97} & 0.89 & 0.00 & 0.71 & 0.79 & \textbf{0.80} & 0.00 & 0.63 & \textbf{0.79} & 0.78 \\
screwdriver & 0.00 & 0.11 & 0.94 & \textbf{0.94} & 0.00 & 0.11 & \textbf{0.90} & 0.48 & 0.00 & 0.11 & \textbf{0.92} & 0.63 & 0.00 & 0.75 & \textbf{0.76} & 0.75 & 0.00 & 0.75 & \textbf{0.76} & 0.74 \\
 \hline Untextured & 0.32 & 0.09 & \textbf{0.95} & 0.86 & 0.09 & 0.09 & \textbf{0.94} & 0.85 & 0.12 & 0.09 & \textbf{0.94} & 0.84 & 0.37 & 0.35 & \textbf{0.81} & 0.79 & 0.37 & 0.24 & \textbf{0.80} & 0.74 \\ \hline
battery\_black & 0.59 & 0.91 & \textbf{0.98} & 0.91 & 0.58 & 0.91 & \textbf{0.99} & 0.97 & 0.58 & 0.91 & \textbf{0.98} & 0.94 & 0.74 & 0.61 & \textbf{0.84} & 0.81 & 0.74 & 0.24 & \textbf{0.84} & 0.81 \\
battery\_green & 0.67 & 0.04 & \textbf{0.94} & 0.84 & \textbf{0.90} & 0.04 & 0.87 & 0.50 & 0.76 & 0.04 & \textbf{0.90} & 0.63 & 0.68 & 0.62 & \textbf{0.80} & 0.75 & 0.68 & 0.54 & \textbf{0.75} & 0.71 \\
box\_brown & 0.37 & 0.24 & \textbf{0.78} & 0.74 & \textbf{0.99} & 0.24 & 0.84 & 0.46 & 0.54 & 0.24 & \textbf{0.81} & 0.57 & \textbf{0.80} & 0.70 & 0.74 & 0.69 & \textbf{0.80} & 0.69 & 0.51 & 0.23 \\
box\_yellow & 0.47 & \textbf{1.00} & 0.96 & 0.96 & 0.82 & \textbf{1.00} & 0.96 & 0.97 & 0.60 & \textbf{1.00} & 0.96 & 0.97 & \textbf{0.85} & 0.84 & 0.84 & 0.83 & \textbf{0.85} & 0.84 & 0.84 & 0.83 \\
glue & 0.59 & 0.94 & \textbf{0.99} & 0.94 & 0.81 & 0.94 & \textbf{0.95} & 0.79 & 0.68 & 0.94 & \textbf{0.97} & 0.86 & 0.78 & 0.79 & \textbf{0.84} & 0.79 & 0.78 & 0.79 & \textbf{0.84} & 0.79 \\
pendrive & 0.88 & 0.81 & \textbf{0.96} & 0.90 & 0.09 & 0.81 & \textbf{0.85} & 0.81 & 0.17 & 0.81 & \textbf{0.90} & 0.85 & 0.64 & 0.63 & \textbf{0.78} & 0.73 & 0.64 & 0.55 & \textbf{0.78} & 0.71 \\
 \hline Textured & 0.58 & 0.58 & \textbf{0.93} & 0.88 & 0.67 & 0.57 & \textbf{0.90} & 0.72 & 0.53 & 0.57 & \textbf{0.91} & 0.78 & 0.75 & 0.72 & \textbf{0.80} & 0.76 & \textbf{0.75} & 0.67 & 0.75 & 0.67 \\ \hline
global & 0.53 & 0.42 & \textbf{0.94} & 0.87 & 0.39 & 0.37 & \textbf{0.92} & 0.80 & 0.45 & 0.40 & \textbf{0.93} & 0.83 & 0.77 & 0.72 & \textbf{0.81} & 0.78 & 0.77 & 0.62 & \textbf{0.78} & 0.73 \\

\bottomrule \end{tabular}

	}
	\vspace{0cm}
	\caption{Performances of our approach, compared to SIFT\cite{lowe1999object}  and BOLD\cite{tombari2013bold}, in the \emph{Hard Scene} scenario (boldface text highlights the best score in the related pane). Here, the overall Precision/Recall index of \algoname{} is about 94\%/92\% showing its robustness against high clutter backgrounds, a typical situation in real industrial environments.}
	\label{tab:results_hard}
	
\vspace*{-6mm}\end{table*}

\subsection{Real Robotic application}\label{sec:qualitative}

%
%
%
%

As a qualitative evaluation of our approach we designed a proof-of-concept pick\&place robotic application based on the outcome of the \algoname{} detector. The setup consist in an industrial robotic arm (COMAU smart six) with a parallel gripper as end effector and an eye-on-hand camera on board. The camera mounted in this way can be thought as the secondary end-effector and than can be arbitrarily moved with high precision. The image plane of the camera is kept parallel to a table-top scene with randomly arranged objects belonging to the \datasetName{} (same camera-table configuration seen in \autoref{sec:dataset_generation}). The distance between camera and the table plane is known. Given that the output of the pipeline is a set of oriented prediction $\breve{y}_i=\{ \breve{b}_i = \{x_i,y_i,w_i,h_i\},\theta_i, c_i \}$ we exploit their position ($x_i,y_i$) and orientation ($\theta_i$) to build a simple control scheme, for robot guidance, in such a way as to move the camera in order to align a target object with the center of camera viewpoint, oriented as the canonical $x$ axis of the image.
A proportional-only control scheme is used, thus, to minimize $\mathbf{e}_t$ and $e_{\theta}$, the translational and rotational error respectively $\mathbf{e}_t = \{ x_i - c_x, y_i - c_y \}$ and $e_{\theta} = -\text{sign}(\sin(\theta_i))(\cos(\theta_i) - 1)$, with $(c_x,c_y)$ the center of the image. The control scheme will produce a linear velocity ${\mathbf{v} = K_{P_t} \mathbf{e}_v}$ and an angular velocity ${\omega = K_{P_\theta} e_{\theta}}$ to the end effector, \ie the camera reference frame ($K_{P_t}$ and $K_{P_\theta}$ are tunable proportional gains). The $\text{sign}(\cdot)$ function is simply the \emph{sign} function with $\text{sign}(0)=1$ to avoid singularities. The above simplified robot guidance scheme is used to lead the robot in a easier condition suitable to estimate the 3D pose of the object because: knowing the extrinsics parameters of the mounted camera and the height of the table w.r.t. the robot base, it is trivial to know the 3D coordinates of an object centered in the camera viewpoint (an alternative solution to estimate 3D coordinates from multiple 2D CNN predictions can be found in \cite{degregorio2018}). \autoref{fig:grasping} exemplifies the described procedure by presenting a real execution of it. In the supplementary material several runs of experiment are shown from the on-board and off-board cameras. We have accomplished this task on 20 scenes with completely unseen backgrounds. 
The overall success rate of the final grasp is $100\%$ for each object, except for \emph{artifcat\_orange} and \emph{box\_brown}, which instead scored $60\%$ and $50\%$ respectively. Hance, the success rate for the entire dataset is $92.5\%$. 
As expected from the conclusions of \autoref{sec:veterans_vs}, the grasp for highly symmetrical objects (like  \emph{artifcat\_orange} and \emph{box\_brown}) becomes very challenging, due to the difficulty of estimating unambiguous orientation for these samples. In these cases, the indecision of the detector combined with the stateless nature of a CNN-based Object Detector (\ie there is no online tracking method like in \cite{milan2017online}, where a Recurrent Neural Network is used in order to achieve a continuous estimate of the pose) leads to a detrimental oscillation of the output in the control module.
Then, we corrected the model by modifying the rotational error in such a way as to treat objects as symmetric, with

\begin{equation}
\begin{cases}
\mathbf{e}_t = \{ x_i - c_x, y_i - c_y \} \\
e_{\theta}^\text{Sym} = \text{sign}(\tan(\theta))(\lvert \sin(\theta) \rvert)
\end{cases}.
\end{equation}

\noindent Accordingly, the robot guidance scheme leads the camera to move towards the nearest horizontal configuration of the target object ($0^{\circ}$ or $180^{\circ}$ indifferently), reaching $100\%$ of success rate also for each object (the control schemes were tested again over 20 scenes). 


\begin{figure}[t]
  \centering
  \includegraphics[width=0.45\textwidth]{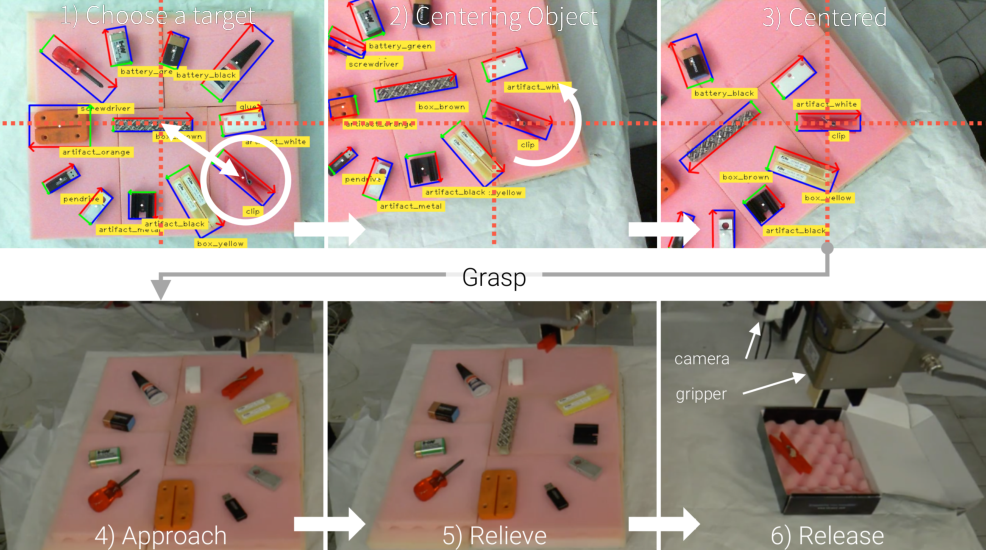}
  \caption{An exemplification of the pick-and-place system designed using \algoname{} as a module of the control scheme. The first row depicts frames coming from the on-board camera, with superimposed predictions, showing the robot guidance control phases. The second row shows images coming from an off-board camera framing the grasp sequence downstream of the alignment procedure. }\label{fig:grasping}
  
\vspace*{-3mm}\end{figure}

\section{CONCLUDING REMARKS}

We proposed an extension of classical CNN-based Object Detectors able to produce Oriented Bounding Boxes suitable for the 3-DoF pose estimation task. We provided our detector with a simplified procedure to gather a huge amount of training data in the field, with trifling human intervention.
With this work we showed how it is possible, with the proposed techniques, to effectively use Deep Learning in a real industrial setting, exploiting a neural network as a module of a more complex control scheme.
We are already working to extend our approach in order to deal also with occlusions ans symmetries together with a set of more complex discretization functions beyond the simple division into equal bins. An interesting future development could be to remove completely human intervention in the dataset generation stage in order to deliver a completely automated, and reliable, industrial system. 




%

\bibliographystyle{IEEEtran}
\bibliography{IEEEabrv,bibliography}

%








\end{document}